\documentclass[letterpaper,twocolumn,fleqn]{article} 

\usepackage{ist}
\usepackage[utf8]{inputenc} 
\usepackage[T1]{fontenc}    
\usepackage{hyperref}       
\usepackage{url}            
\usepackage{booktabs}       
\usepackage{amsfonts}       
\usepackage{nicefrac}       
\usepackage{microtype}      
\usepackage{graphicx} 
\usepackage{multicol}[1999/05/25]
\usepackage{blindtext}
\usepackage[T1,T5]{fontenc}
\title{Gun Source and Muzzle Head Detection}
\author{%
Zhong Zhou\textsuperscript{1}, Isak Czeresnia Etinger\textsuperscript{1}, Florian Metze\textsuperscript{1}, Alexander Hauptmann\textsuperscript{1}, Alexander Waibel\textsuperscript{1, 2}\\[1ex]
1 Carnegie Mellon University, Pittsburgh, PA, the USA \\
2 Karlsruhe Institute of Technology, Karlsruhe, Germany }

\date{} 
\begin{document}
\maketitle 
\thispagestyle{empty} 

\begin{abstract}
 There is a surging need across the world for protection against gun violence. There are three main areas that we have identified as challenging in research that tries to curb gun violence: temporal location of gunshots, gun type prediction and gun source (shooter) detection. Our task is gun source detection and muzzle head detection, where the muzzle head is the round opening of the firing end of the gun. We would like to locate the muzzle head of the gun in the video visually, and identify who has fired the shot.
  In our formulation, we turn the problem of muzzle head detection
  into two sub-problems of human object detection and gun smoke detection. Our assumption
  is that the muzzle head typically lies between the gun smoke caused by the shot and the shooter. We have
  interesting results both in bounding
  the shooter as well as detecting
  the gun smoke. In our experiments, we are successful in detecting the muzzle head by detecting the gun smoke and
  the shooter. 
\end{abstract}

\section{Introduction} 
There is a surging need across the world for protection against
gun violence. There are 17,502 gun violence incidents that resulted 
in 4606 deaths in the United States alone up to date; among which 105 are 
mass shooting as shown in the Gun Violence Archive 
\footnote{\url{https://www.gunviolencearchive.org}}. 
Mass shooting is defined by the Mass Shooting Tracker 
\footnote{\url{https://www.shootingtracker.com}} as any incident that involves shooting 4 people or more in one incident independent 
of any circumstance. 

From the 2017 Las Vegas shooting where rows of semi-automatic machine guns 
were pointed at innocent people, to the 2017 Texas Sutherland church shooting 
where many were killed during Sunday worship; from the 2016 nightclub 
shooting where many members of the gay community were killed, to the 2018 
Pittsburgh Tree of Life synagogue shooting where a lot of animosity were shown towards 
innocent members of the Jewish community. The loss of life, the contribution 
to group segregation, tension, division and conflict, 
and the damage done to the survivors and to
each community targeted is immeasurable. Therefore, 
there is an insurmountable and tremendous need for 
curbing gun violence for humanitarian efforts \footnote{Images of various shooting events in this paper are taken from \url{https://www.wcjb.com}, \url{https://www.chabad.org}, \url{https://www.nbcnews.com}, and \url{https://www.bbc.com/}}. 
\begin{figure}[t]
  \center
  \includegraphics[scale=0.27]{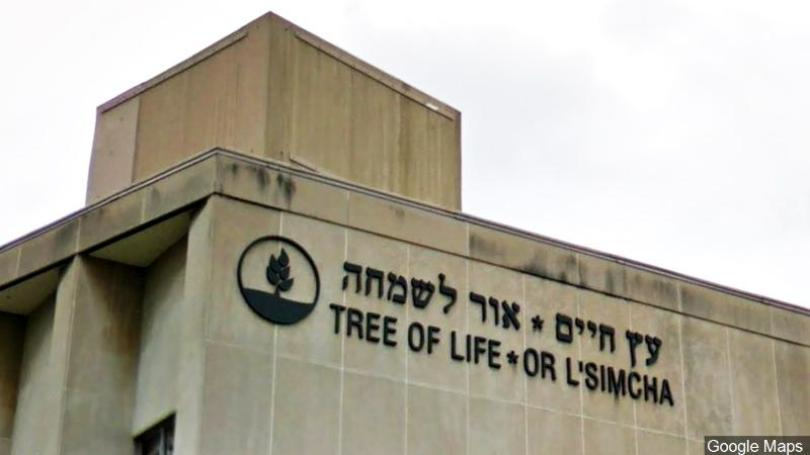}
  \caption{Tree of Life synagogue \cite{ToLS}.}
  \label{fig:treeoflife}
\end{figure}
\begin{figure*}[h!]
\centering
\includegraphics[scale=0.8]{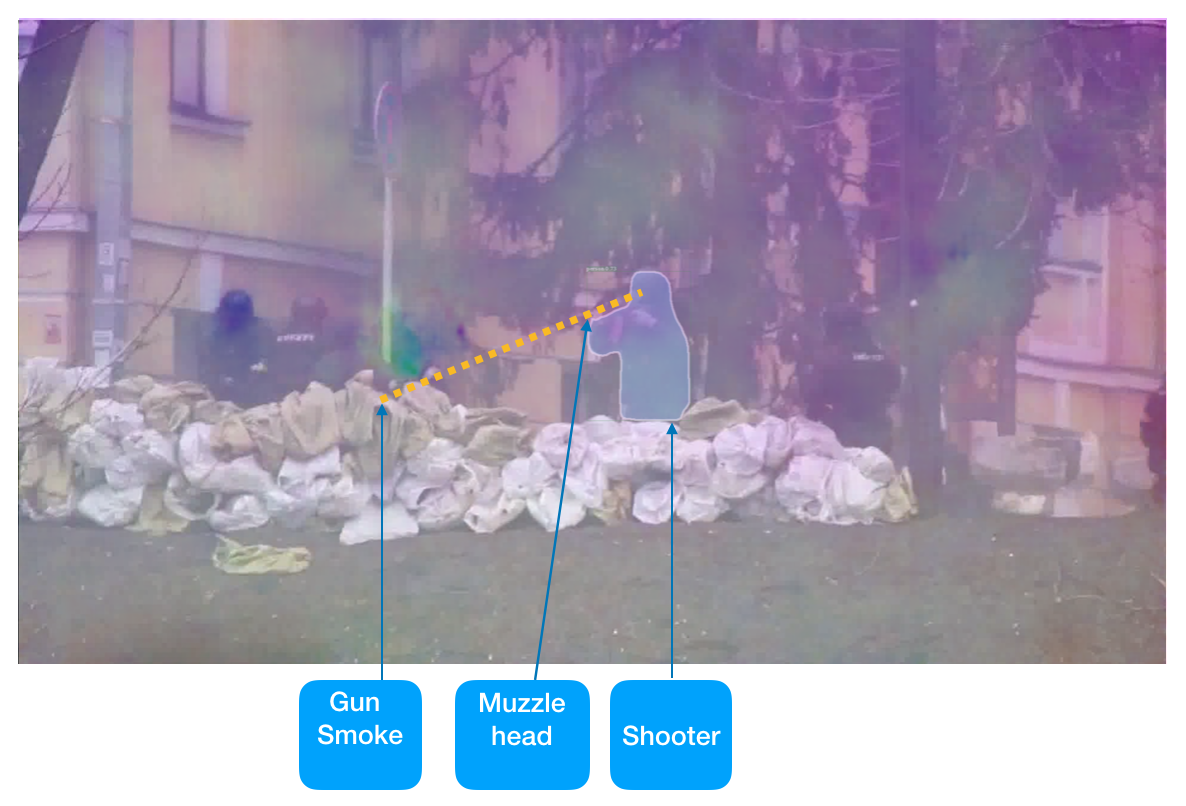}
\caption{Overlay of optical flow visualization over the original video with human detection \cite{he2017mask}.} 
\label{fig:final}
\end{figure*}

Many researchers are interested in contributing to the efforts to mitigate 
gun violence. There are three main areas that we have identified as 
challenging: temporal location of gunshots, gun type prediction and
source detection. Temporal location of gunshots in essence is finding the probability distribution of the likelihood of a gunshot
to be found in a temporal segment of the audio. It is very important to 
extract accurate signal at this stage for subsequent research. 
Gun type detection involves detection and discerning 
which type of gun is involved. Gun source detection involves 
finding out who fires the shot, especially in the scenario 
where there are multiple people in the video frame. 
And there may be multiple people carrying guns in the video frame but not 
everyone is shooting. There are many researchers who focused on 
the first two problems and achieved good results through Localized 
Self-Paced Reranking by 
\cite{liang2017temporal, ruonan2019gun, tropp2007signal, chen2001atomic, coifman1992wavelet, zhang2015human, ankit2019gun}.

Our task is gun source detection and muzzle head detection. 
Muzzle head is the round opening of the firing end of the gun. 
We would like to visually locate the muzzle head of 
the gun in the video, and identify who has fired the shot. This is a very difficult 
task. There are very few real-life events that contain the footage 
of the shooter, compared to a multitude of videos of the victims. 
Even in situations where both gun and shooter are visible, it is difficult
to visually detect whether there is indeed a gunshot, it is therefore 
even more difficult to discern who fires it. 
It usually takes a long time to manually detect who fires the shot, and 
in many cases, even manual detection would fail. There is a huge 
gain in automating the process, or at least in greatly facilitating human efforts on detecting who fires the shot 
and locate the muzzle head where a shot is fired.

\section{Related Work}
Many researchers have worked on gunshot prediction as well
as gun type detection using audio data. The first wave of work with this strategy
was performed between 2005 and 2013.

 Researchers detected abnormal audio events in continuous audio recordings of public places and focused on robustness and prioritized recall over precision \cite{clavel2005events}. Some worked on the modelling of gunshot trajectories by simple geometry and kinematics, using the time taken for sound to travel from a gun to a recorder as an indication of distance from the recorder \cite{maher2006modeling}. Some expanded on the work of the two previous papers by building a system using two Gaussian Mixture Model classifiers for detecting gunshots and screams, and also using kinematics to model gunshot trajectories \cite{valenzise2007scream, gerosa2007scream}. Some used dynamic programming and Bayesian networks to detect gunshots from audio streams from movies \cite{pikrakis2008gunshot}. Many extended evaluation of previous techniques to extremely noisy environments, by recording gunshots at open fields and adding white noise \cite{freire2010gunshot}. The efficiency of several methods from 2005 to 2010 is summarized by \cite{chacon2011evaluation}.

\begin{figure}[t]
  \center
  \includegraphics[scale=0.32]{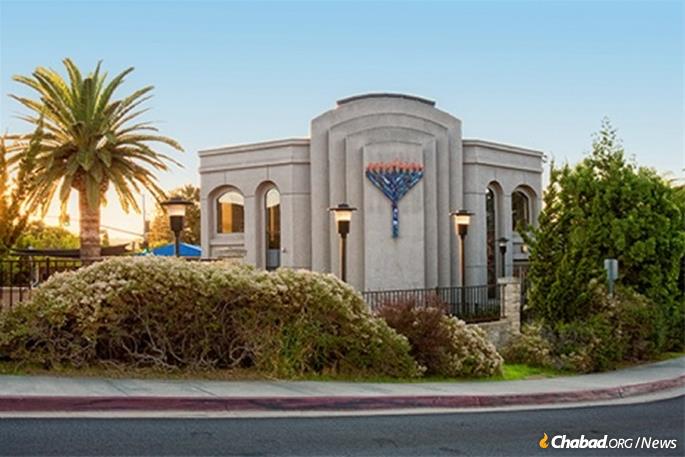}
  \caption{Chabad of Poway Synagogue (near San Diego) \cite{CoPS}.}
  \label{fig:ChabadPoway}
\end{figure}
\begin{figure}[b]
  \center
  \includegraphics[scale=0.14]{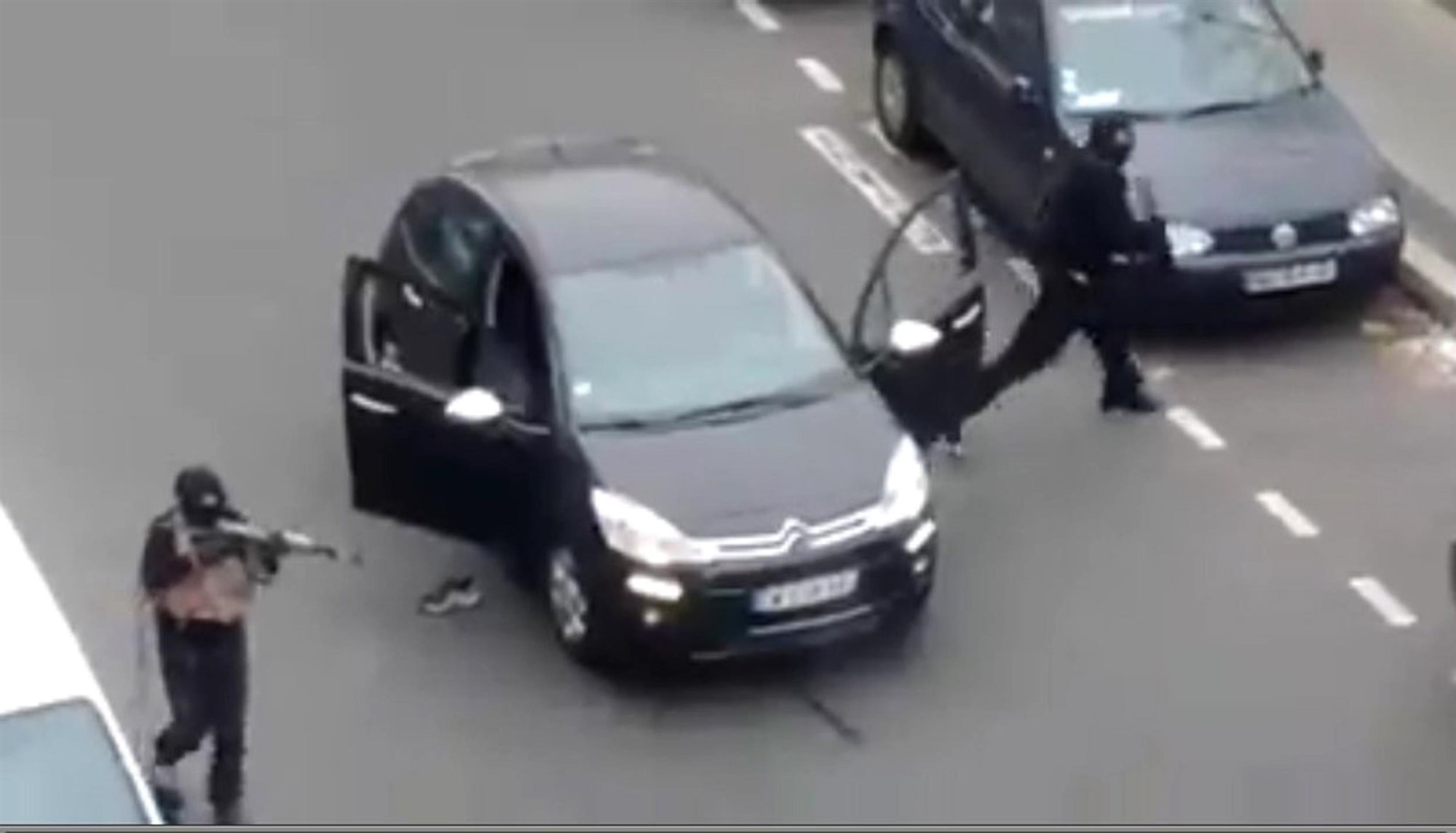}
  \caption{Charlie Hebdo terrorist shooting \cite{CHts}.}
  \label{fig:CharlieHebdo}
\end{figure}
\begin{figure}[t]
  \center
  \includegraphics[scale=0.25]{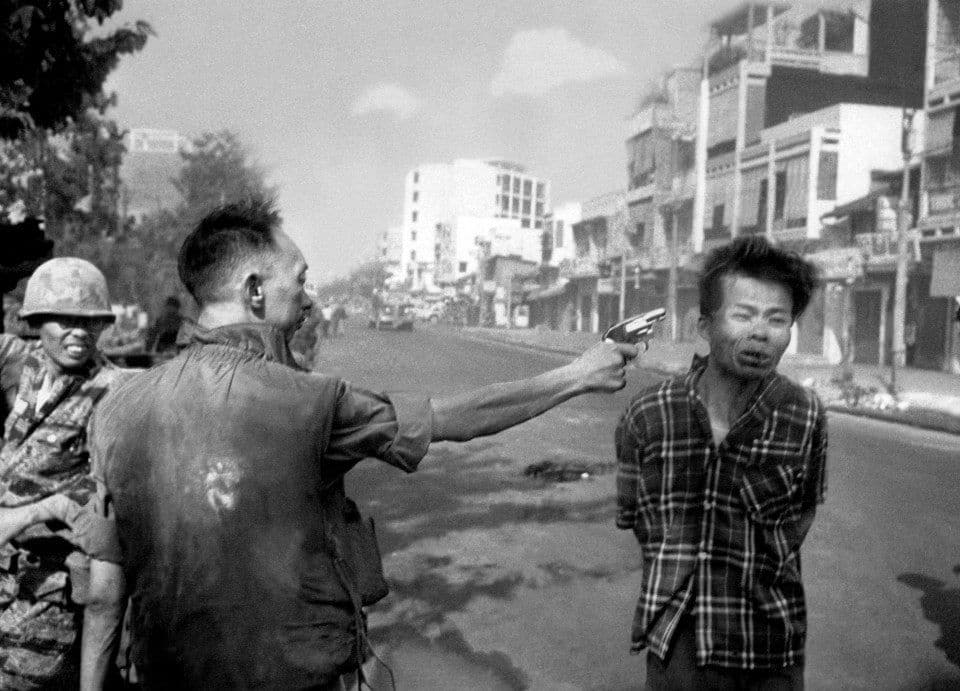}
  \caption{Execution of {\fontencoding{T5}\selectfont Nguy\'\ecircumflex{}n V{\abreve}n L\'em}, by Eddie Adams.}
  \label{fig:example}
\end{figure}

\begin{table*}[h!]
	\centering
    \tabcolsep=0.1cm
	\begin{tabular}{|c|c|c|c|c|c|c|c|c|c|c|}
		\hline
		\begin{tabular}{@{}c@{}}\rmfamily{Video} \\ \rmfamily{id}\end{tabular} & \begin{tabular}{@{}c@{}}\rmfamily{Smoke color+} \\ \rmfamily{intensity[1--5]}\end{tabular} & \begin{tabular}{@{}c@{}}\rmfamily{Background} \\ \rmfamily{color}\end{tabular} & 
		\begin{tabular}{@{}c@{}}\rmfamily{Video} \\ \rmfamily{resolution}\end{tabular} & 
		\begin{tabular}{@{}c@{}}\rmfamily{Camera} \\ \rmfamily{far?}\end{tabular} & 
        \begin{tabular}{@{}c@{}}\rmfamily{Gun} \\ \rmfamily{stable?}\end{tabular} & 
        \begin{tabular}{@{}c@{}}\rmfamily{Shooter} \\ \rmfamily{moves?}\end{tabular} & 
        \begin{tabular}{@{}c@{}}\rmfamily{Camera} \\ \rmfamily{moves?}\end{tabular} &
        \begin{tabular}{@{}c@{}}\rmfamily{Gun position} \\ \rmfamily{w.r.t. camera}\end{tabular} &
        \begin{tabular}{@{}c@{}}\rmfamily{Shot/shooter} \\ \rmfamily{obstruction}\end{tabular} &
        \begin{tabular}{@{}c@{}}\rmfamily{Shooter} \\ \rmfamily{pose}\end{tabular}
        \\
        \hline
		\rmfamily{1} & \rmfamily{grey, 5} & \rmfamily{grey} & \rmfamily{good} & \rmfamily{no} & \rmfamily{no}  & \rmfamily{no} & \rmfamily{no} & \rmfamily{pointed up} & \rmfamily{people} & \rmfamily{standing} \\
		\rmfamily{2} & \rmfamily{grey, 2} & \rmfamily{grey} & \rmfamily{medium} & \rmfamily{no} & \rmfamily{yes} & \rmfamily{no} & \rmfamily{no} & \rmfamily{sideways} & \rmfamily{nothing} & \rmfamily{standing} \\
		\rmfamily{3} & \rmfamily{grey, 1} & \rmfamily{grey} & \rmfamily{bad} & \rmfamily{no} & \rmfamily{yes} & \rmfamily{no} & \rmfamily{no} & \rmfamily{sideways} & \rmfamily{nothing} & \rmfamily{kneeling} \\
		\rmfamily{4} & \rmfamily{orange, 5} & \rmfamily{white} & \rmfamily{bad} & \rmfamily{no} & \rmfamily{no} & \rmfamily{yes} & \rmfamily{no} & \rmfamily{sideways} & \rmfamily{nothing} & \rmfamily{kneeling} \\
		\rmfamily{5} & \rmfamily{grey, 1} & \rmfamily{white} & \rmfamily{medium} & \rmfamily{no} & \rmfamily{yes} & \rmfamily{no} & \rmfamily{no} & \rmfamily{sideways} & \rmfamily{nothing} & \rmfamily{lying} \\
		\rmfamily{6} & \rmfamily{orange, 5} & \rmfamily{white} & \rmfamily{bad} & \rmfamily{no} & \rmfamily{no} & \rmfamily{yes} & \rmfamily{no} & \rmfamily{behind} & \rmfamily{nothing} & \rmfamily{standing} \\
		\rmfamily{7} & \rmfamily{grey, 2} & \rmfamily{grey} & \rmfamily{medium} & \rmfamily{no} & \rmfamily{yes} & \rmfamily{no} & \rmfamily{no} & \rmfamily{sideways} & \rmfamily{nothing} & \rmfamily{standing} \\
		\rmfamily{8} & \rmfamily{grey, 1} & \rmfamily{grey} & \rmfamily{medium} & \rmfamily{no} & \rmfamily{yes} & \rmfamily{no} & \rmfamily{no} & \rmfamily{sideways} & \rmfamily{nothing} & \rmfamily{lying} \\
		\rmfamily{9} & \rmfamily{grey, 1} & \rmfamily{grey} & \rmfamily{medium} & \rmfamily{no} & \rmfamily{no} & \rmfamily{yes} & \rmfamily{no} & \rmfamily{sideways} & \rmfamily{tree} & \rmfamily{walking} \\
		\rmfamily{10} & \rmfamily{grey, 1} & \rmfamily{grey} & \rmfamily{medium} & \rmfamily{no} & \rmfamily{no} & \rmfamily{yes} & \rmfamily{no} & \rmfamily{sideways} & \rmfamily{tree} & \rmfamily{walking} \\
		\rmfamily{11} & \rmfamily{orange, 1} & \rmfamily{grey} & \rmfamily{medium} & \rmfamily{no} & \rmfamily{yes} & \rmfamily{no} & \rmfamily{no} & \rmfamily{sideways} & \rmfamily{nothing} & \rmfamily{kneeling} \\
		\rmfamily{12} & \rmfamily{orange, 4} & \rmfamily{grey} & \rmfamily{medium} & \rmfamily{no} & \rmfamily{yes} & \rmfamily{no} & \rmfamily{no} & \rmfamily{sideways} & \rmfamily{nothing} & \rmfamily{kneeling} \\
		\rmfamily{13} & \rmfamily{grey, 1} & \rmfamily{grey} & \rmfamily{bad} & \rmfamily{yes} & \rmfamily{yes} & \rmfamily{no} & \rmfamily{no} & \rmfamily{sideways} & \rmfamily{nothing} & \rmfamily{standing} \\
		\rmfamily{14} & \rmfamily{grey, 2} & \rmfamily{grey} & \rmfamily{good} & \rmfamily{no} & \rmfamily{yes} & \rmfamily{no} & \rmfamily{yes} & \rmfamily{sideways} & \rmfamily{tree} & \rmfamily{standing} \\
		\rmfamily{15} & \rmfamily{grey, 3} & \rmfamily{grey} & \rmfamily{medium} & \rmfamily{no} & \rmfamily{no} & \rmfamily{yes} & \rmfamily{yes} & \rmfamily{sideways} & \rmfamily{nothing} & \rmfamily{walking} \\
		\hline
	\end{tabular}
	\label{tbl:Stats1}
    \caption{Statistic of videos with visible gunshots.}
\end{table*}

\begin{figure*}[h!]
\centering
\includegraphics[height=140mm]{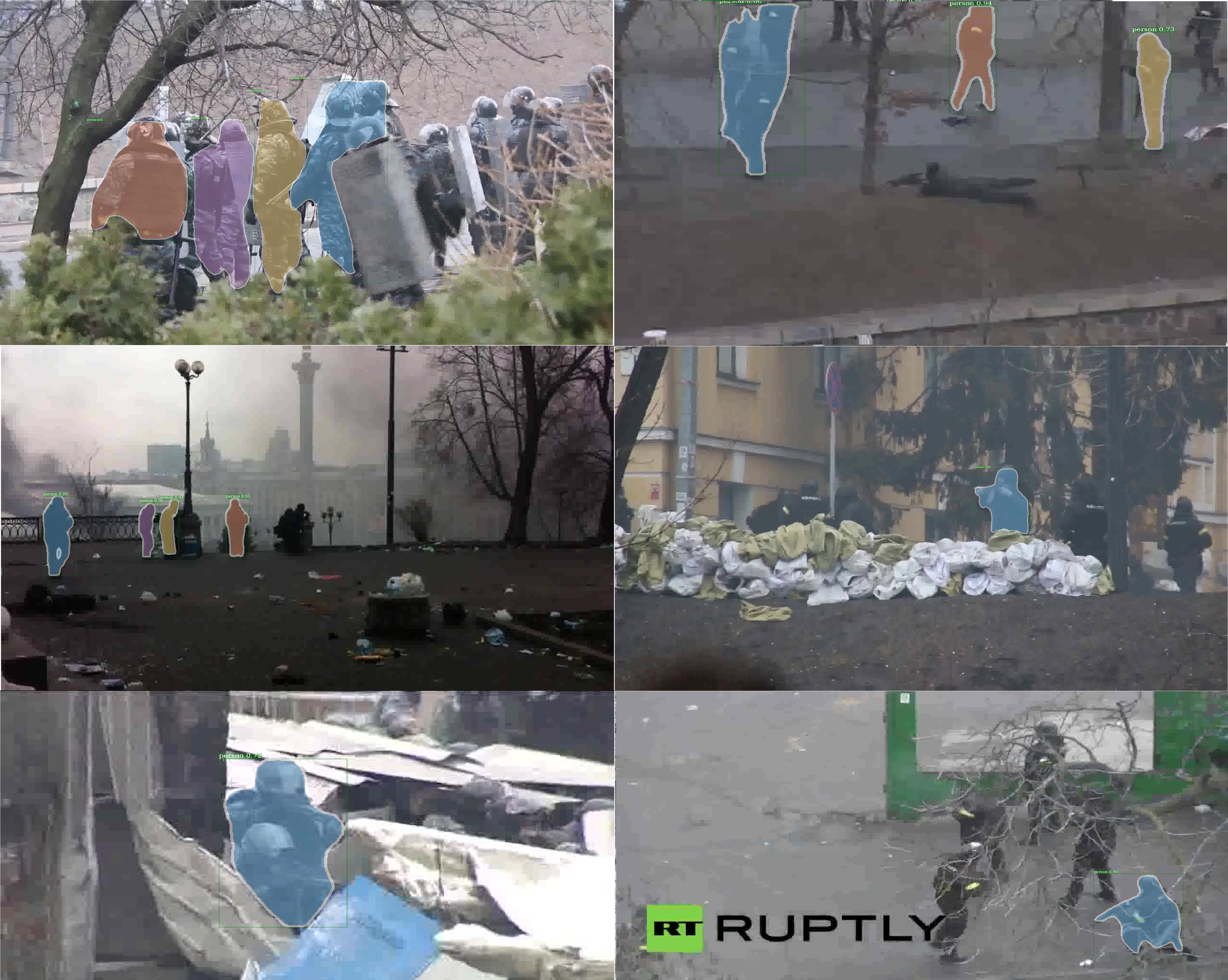}
\caption{Results from human object detection mechanism through 
the Detectron network \cite{Detectron2018}.} 
\label{fig:people_merged}
\end{figure*}
\begin{figure*}[h!]
\centering
\includegraphics[height=180mm]{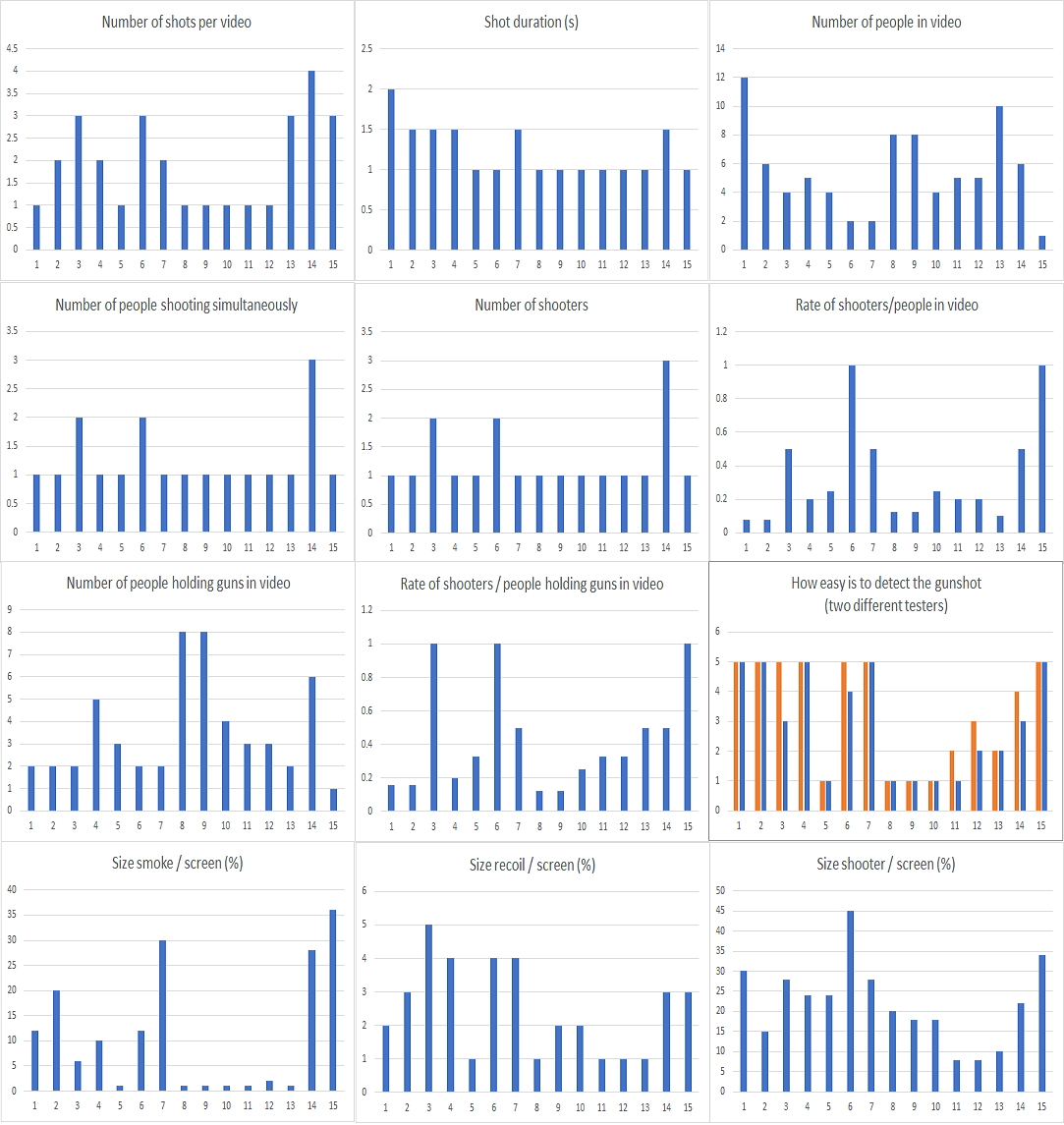}
\caption{Statistics of videos with visible guns. The horizontal axis is the video id.}
\label{fig:stats2}
\end{figure*}

\begin{figure}[h!]
\centering
\includegraphics[scale=0.68]{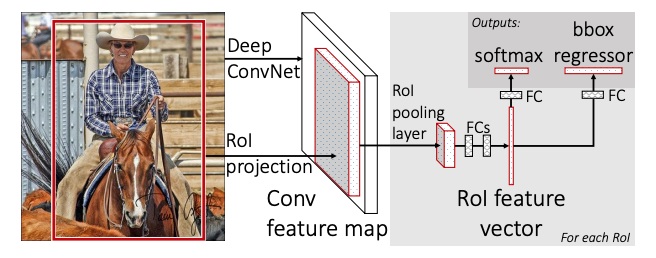}
\caption{fast R-CNN architecture \cite{girshick2015fast}.}
\label{fig:fastRCNN}
\end{figure}
\begin{figure}[h!]
\centering
\hspace*{-0.7cm}
\includegraphics[scale=0.5]{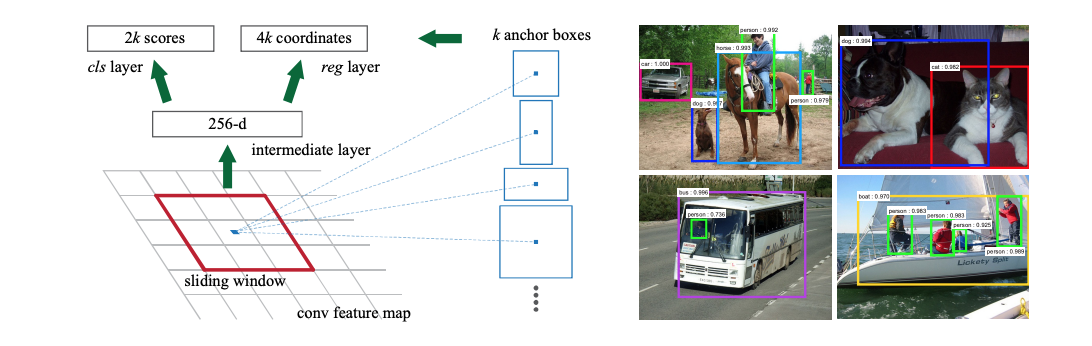}
\caption{Faster R-CNN's use of RPN \cite{ren2015faster}.}
\label{fig:fasterRCNN_RPN}
\end{figure}
\begin{figure}[h!]
\centering
\includegraphics[scale=0.68]{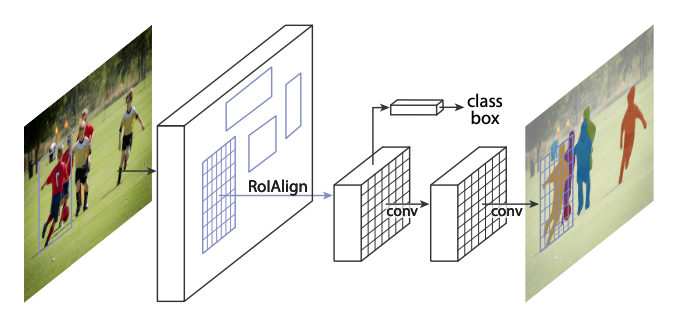}
\caption{Mask R-CNN architecture \cite{he2017mask}.} 
\label{fig:MaskRCNN}
\end{figure}
\begin{table}[b]
	\centering
    \begin{tabular}{ |p{2.5cm}|p{2.5cm}|p{2.5cm}| }
    \hline
    \rmfamily{Gun Cloud Detection Success rate} & \rmfamily{Shooter Detection Rate} & \rmfamily{Muzzle Head Detection Rate} \\ \hline\hline
    \rmfamily{69.6}\% & \rmfamily{30.4}\% & \rmfamily{21.7}\%  \\ \hline 
    \end{tabular}
    \label{tbl:final}
    \caption{\rmfamily{Performance of gun smoke, shooter, muzzle head Detection.}}
\end{table}

\begin{figure}[b]
  \center
  \includegraphics[scale=0.15]{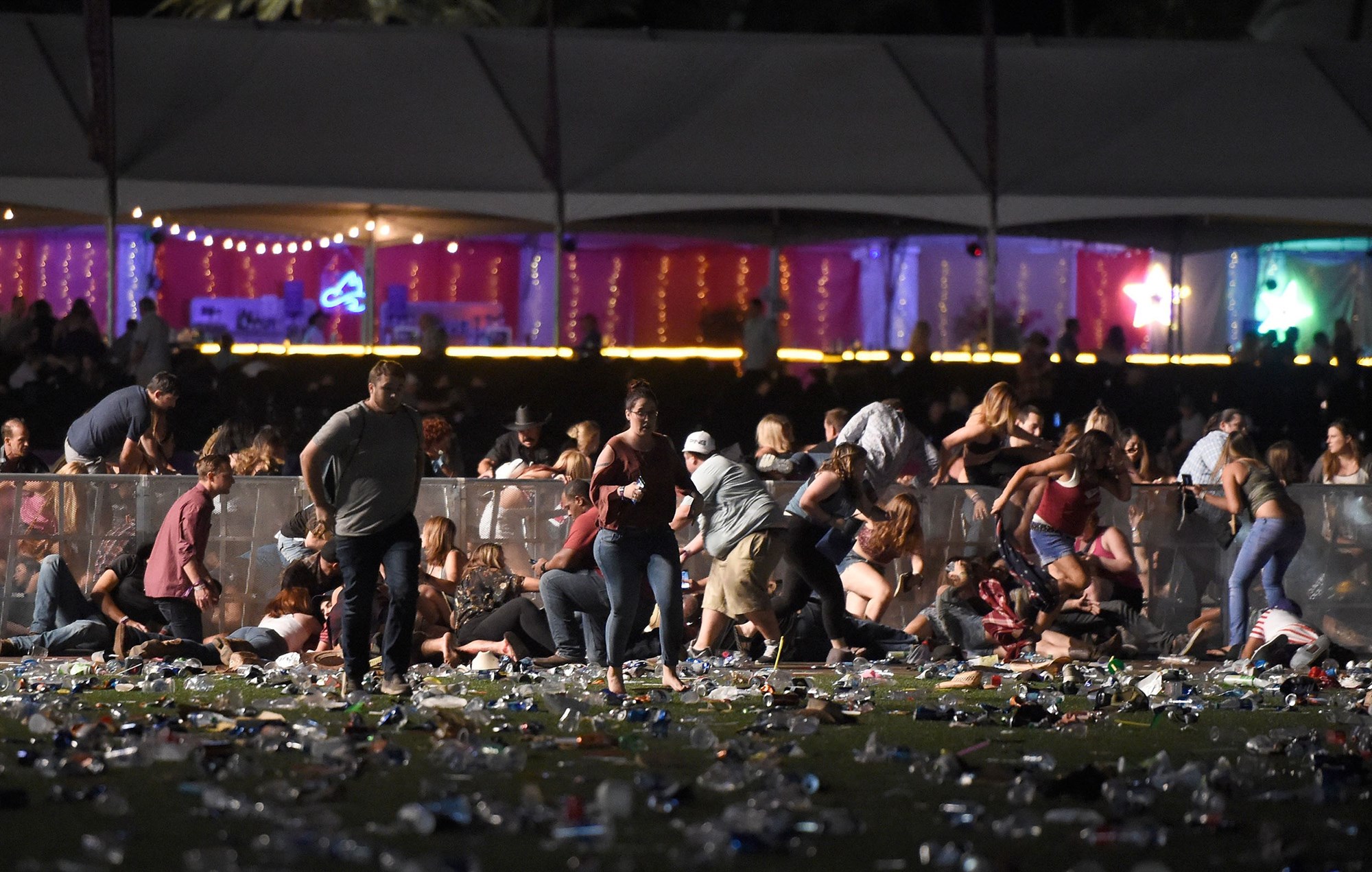}
  \caption{Las Vegas shooting \cite{LVs}.}
  \label{fig:lasVegas}
\end{figure}
\begin{figure*}[h!]
\centering
\includegraphics[height=140mm]{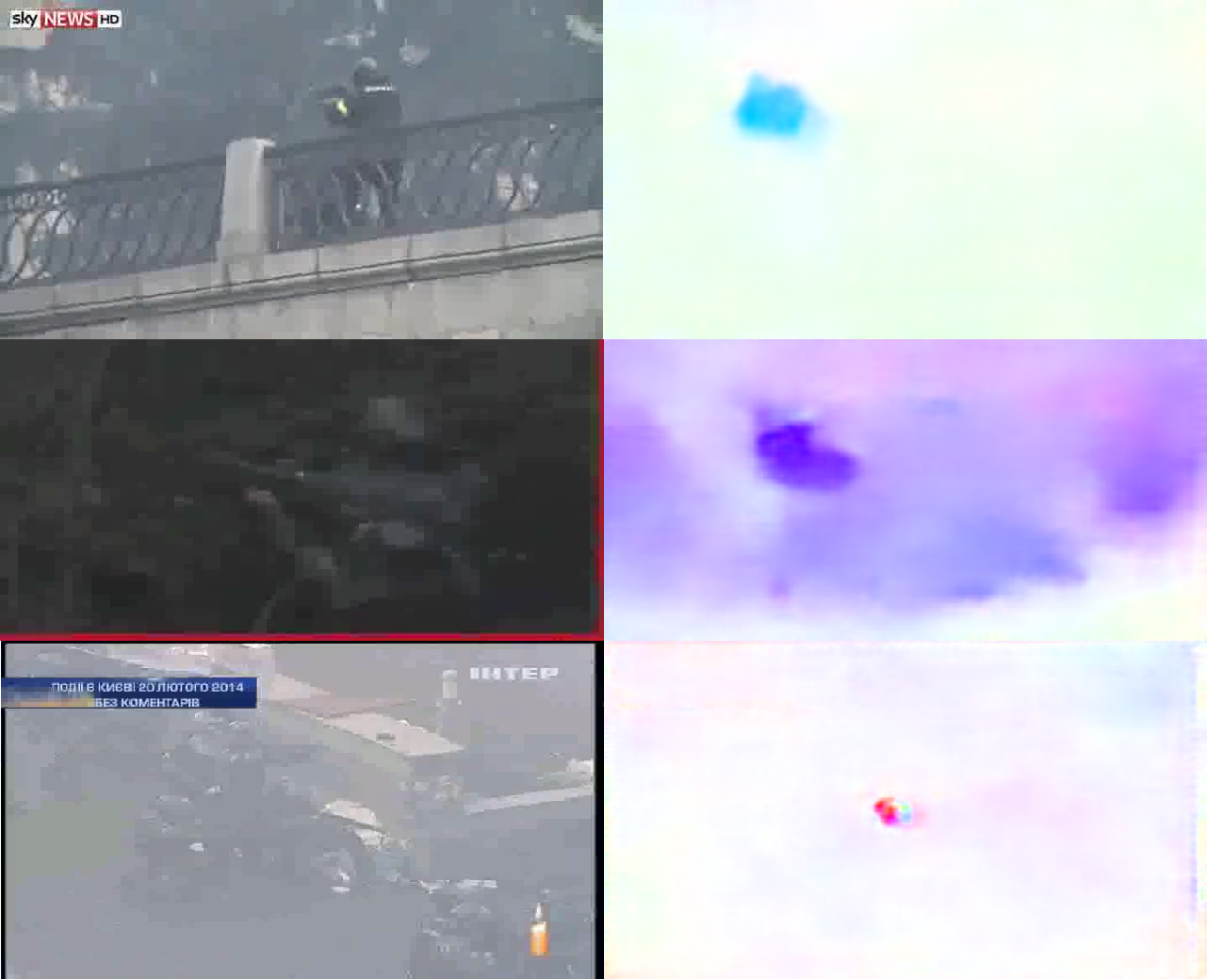}
\caption{Comparison between original video frame and its respective Optical Flow graph. Left: Video keyframe with Visible Shooting. Right: Optical Flow Graph showing clearly the shooting smoke.}
\label{fig:bubble_merged}
\end{figure*}

In 2012, Kumar's team published a work that automatically learned atomic units of sound, called Acoustic Unit Descriptors \cite{kumar2012audio}. In 2013, Ahmed formulated a two-stage approach that uses an event detection framework followed by a gunshot recognition stage, which used a template matching measure and the eighth order linear predictive coding coefficients to train an SVM classifier \cite{ahmed2013improving}.

A second wave of more advanced methods came later \cite{liang2016video, liang2017temporal, lim2017rare, ruonan2019gun}. Liang created a tool that synchronizes multiple videos of a single event (especially event involving gunshots) by creating a unique sound signature at the frame level for each video in a collection \cite{liang2016video}.
Liang worked on the temporal localization of gunshots, and employed Localized Self-Paced Reranking (LSPaR) to refine the localization results \cite{liang2017temporal, jiang2014easy}. LSPaR utilizes curriculum learning (i.e.: samples are fed to the model from easier to noisier ones) so that it can overcome the noisiness of the initial retrieval results from user-generated videos. Additionally, SPaR has a concise mathematical objective to optimize and useful properties that can be theoretically verified, instead of relying mainly on heuristic weighting as other reranking methods.
Lim introduces a rare sound event detection system using a 1D convolutional neural network and long short term memory units (LSTM). MFCCs are used as input; the 1D ConvNet is applied in each time-frequency frame to convert the spectral feature; then the LSTM incorporates the temporal dependency of
the extracted features \cite{lim2017rare}.

Audio-based gunshot detection and temporal 
localization have many applications. 
Some 
are patents of real products \cite{baxter2007gunshot, cowdry2014gun}. Aronson  
developed a tool to monitor conflicts \cite{aronson2015video}. 
This is because audio is often the only source 
available. For gun type detection, clean video
gunshots with very explicit source 
like the well-known Eddie Adams' Vietnam 
war picture (Figure \ref{fig:example}) are very rare.
In general terrorist attacks, it is very 
unlikely to have people recording data
because attacks are unsuspected (so people 
would have no reason to be filming the action
beforehand), and because people seek cover 
as soon as the threat is realized. For example, 
Chen showed that the vast 
majority of recorded data of the Boston Marathon
terrorist attack was not directly aimed at the action \cite{chen2016videos}. 

However, with the rapid increase in the use of 
smart phones with high-resolution cameras in the past
several years, there has been a growing 
opportunity to leverage video data for gunshot detection
and localization. As most of the analyzed video 
data of terrorist attacks or human-rights violations
does not usually involve gunshots, it is 
hard to find useful real-life data for research. 
Even in the few videos where gunshots are observed, 
the guns are usually not visible or extremely small, 
this renders the task to be very challenging. Take the well-known Eddie Adams' Vietnam 
war picture for example again, it is very rare that we have 
the shooter, the gun, and the victim in 
the same picture. For the 2017 Las Vegas shooting, thousands 
of videos are on the screaming crowd of victims, but there 
is no footage about the shooter, because the shooter 
is shooting from a highly elevated hotel room using rows of 
semi-automatic machines guns that are invisible from 
any visual sources. Indeed, videos with complete information 
of the guns are very rare. 

There is much research done on image-based 
gun detection \cite{sun2001segmentation, xue2002fusion, zhang1997region, tiwari2015computer}. Zhang was one of the pioneers in image-based gun detection. He worked on region-based image fusion for concealed weapon detection \cite{zhang1997region}.
Sun then worked on the detection of gun barrels by using segmentation of forward-looking infrared (FLIR) images with fuzzy thresholding and edge detection \cite{sun2001segmentation}.
Xue compared the performances of a large set of image fusion algorithms for concealed weapon detection using visual and infrared images \cite{xue2002fusion}.
Finally, Tiwari proposed a framework that exploits the color based segmentation to eliminate unrelated object from an image using k-means clustering, then Harris interest point detector and Fast Retina Keypoint is used to locate the guns in the segmented images \cite{tiwari2015computer}.

But still, this is done using only images -- not 
complete video data. To the best of our knowledge, there is 
no existing literature that uses video data
as well as audio data together. The temporal 
localization of gunshots and the gun type detection
are very well done, mainly based purely on sound 
\cite{liang2017temporal, lim2017rare, ruonan2019gun, ankit2019gun}. 

However, gunshot source detection is still 
very untackled, and this work seems to leverage
the combination of sound and video to address that challenge. We differentiate from previous research not only by using both visual and sound data jointly, but also by using video features instead of only static image features.

\section{Data}
We have four datasets: Urban Conflict, Target Range, Urban Sound, and Real-World Events.

\subsection{Urban Conflict}
The Urban Conflict dataset consists of 422 videos of conflicts in Ukraine gathered by news sources. The videos are split into 4537 segments, and they sum to 52.5 hours in total duration. The average length of each video is therefore 7.5 minutes.

\subsection{Target Range}
The Target Range dataset consists of 547 soundtracks of gunshots downloaded from YouTube videos of target range practices, further divided into 3761 segments \cite{liang2017temporal}.

\subsection{Urban Sound}
The Urban Sound dataset consists of 1302 audio files of field recordings \cite{salamon2014dataset}. This dataset provides sound on a wide range of noises common to the urban environment. We used only the gunshot data from this dataset. Systematic urban sound classification is a new field of research with many applications. There is scarcity of a common taxonomy. 

\subsection{Real-World Events}
The Real-World Event dataset was gathered from recordings of several different events: Las Vegas Shooting, Santa Fe School shooting, Orlando Night Club shooting, Douglas High School shooting, JacksonVille Tournament shooting, Dallas shooting, Florida school shooting, Thousand Oaks shooting, Sutherland Springs church shooting, Kansas taser shooting \cite{ankit2019gun}. 

\section{Experiments, Methods, and Data Flow}
\subsection{Gunshot Probability Distribution through Localized Self-Paced Reranking}

We applied the model from Liang on the Urban Conflict dataset \cite{liang2017temporal}. The process consists of three steps:
\begin{enumerate}
    \item First the audio stream is extracted from the videos and chunked into small segments of 3-second windows with a 1 second stride. Bag-of-Words of MFCC features is employed to represent each segment \cite{jin2012event, liang2015detecting}.
    \item Then, two-class SVM classifiers are trained for each audio event and applied to the video segments from test videos. However, the initial detection results have low accuracy due to noise.
    \item After the detector model produces an initial ranked list of video segments, we utilize LSPaR to learn a reranking model with curriculum learning (first "cleaner" videos with high confidence scores are fed, then "noisier" videos with smaller confidence scores).
\end{enumerate}

This produced a list of 3-second video segments ranked by a confidence score of how likely there is a gunshot in the video, based purely on audio information. 

\subsection{Manual Selection of Videos with Visible Shooting}
Once we obtained the probability distribution of possible gunshots in each video segment, we choose the video segments with the probability of gunshot (as calculated in the previous Subsection) exceeding a threshold of 70\%. After this step, we manually watched all videos to find the videos that contain visible gunshots.

We found 15 non-overlapping small segments of videos that contained a visible gunshot. Many of those videos were extremely hard to find manually. The reasons are the low resolution of the videos, the small size of the gun recoil and the smoke relative to the screen size;  in addition, there are typically many hundreds of hours of videos to be analyzed before a gunshot is observed. Several statistics of those 15 videos can be found on Table \ref{tbl:Stats1} and in Image \ref{fig:stats2}.

\subsection{Human Object Detection Through Detectron}
After the audio temporal localization of gunshots and 
the manual search for videos with visible shooting 
scenes, we obtained a set of 
15 non-overlapping small segments of videos that contained a visible gunshot, and many had multiple gunshots happening concurrently. 
After examining the videos closely, we realize that the camera is 
moving in many of the videos. As the camera moves, all objects in the frame 
moves. Therefore, it is hard to 
track a fixed point like the muzzle head automatically. 

Our idea is to track the shooter instead 
of tracking the muzzle head. There are several 
reasons for this. Firstly, the shooter is much bigger 
than the muzzle head, instead of doing small (tiny) 
object detection, we can do usual object detection, 
which is researched on more extensively by the 
scientific community. Secondly, there are 
existing open-source models trained on people detection, 
but not on muzzle head because there are very few instances 
of muzzle head in most of the image datasets. 
Thirdly, even though the camera 
may move, the shooter and the muzzle head are relatively 
static with respect to each other as the muzzle head is in the 
direction of the shooter's eyes and is in the direction where 
the shooter is facing. 

In order to detect people in images, we use Facebook's implementation of Mask R-CNN, Detectron \cite{Detectron2018}. Fast R-CNN has achieved huge success in the visual domain  as shown in Figure \ref{fig:fastRCNN} \cite{girshick2015fast}. It has advantages over the state-of-the-art object detection benchmark R-CNN and SPPnet in that it has much better mAP scores (better detection accuracy) while training in the single-stage fashion utilizing multi-task loss function. Every network layer is updated concurrently at the same time during training while no extra disk space is needed during each stage of feature caching. 

Faster R-CNN builds on Fast R-CNN by presenting a Region Proposal Network (RPN) as shown in Figure \ref{fig:fasterRCNN_RPN} \cite{ren2015faster}. RPN enables sharing of CNN features with the network that does object detection facilitating low-cost region proposal scheme. 

Mask RCNN carries Faster R-CNN further by adding an object mask prediction on top of the bounding box prediction
as shown in Figure \ref{fig:MaskRCNN}. We use Facebook's implementation of Mask RCNN to detect people in our videos \cite{Detectron2018}. 

\subsection{Human Evaluation of People Detection}
For most videos, our results of human detection are good; however, when videos have low-resolution or when there are obstructions between the camera man and the shooter(s), it is hard for the human detection mechanism to detect all the shooters involved as shown in Figure \ref{fig:people_merged}. We also find that if a shooter lies flat on the ground, it is hard for detectron to detect this shooter. This is a very interesting problem because 
people, unlike guns or muzzle head, are deformable; a person is capable of having different shooting poses as shown in Table \ref{tbl:Stats1} \cite{moore1999exploiting}. Most of the shooters shoot behind some barriers, either standing or squatting down, and very few training data involves shooters lying flat on the ground. One of the hardest pose of a shooter is the pose of shooting while lying on the ground. In human detection training data, the majority of the data samples involves people standing, sitting, it is extremely sparse to find training data where a person is lying flat on the ground. Therefore, identifying shooters lying on the ground is challenging. 

\subsection{Optical Flow, and Flownet 2.0}
In order to detect visible shooting cloud, we use FlowNet 2.0, 
which is the evolved estimation of Optical Flow algorithm 
using deep Neural Networks \cite{ilg2017flownet}. FlowNet formulated Optical Flow estimation in 
deep Convolutional Networks as a supervised learning problem \cite{dosovitskiy2015flownet}. FlowNet 2.0 enhanced Flownet to cover minor movements and noisy real-world data by adding a small sub-network that covers minor movements and utilizing a stacked network scheme  \cite{ilg2017flownet}. 
We employ flow2image \footnote{Available at \url{https://github.com/georgegach/flow2image}, originally based on Daniel Scharstein (C++) and Deqing Sun (MATLAB)'s work \url{http://vision.middlebury.edu/flow/}} to visualize our outputs from Flownet 2.0. In Figure \ref{fig:bubble_merged}, the right hand side contains the flow visualization. 

\subsection{Human Evaluation of flow visualizations}
In Figure \ref{fig:bubble_merged}, we show a few examples of our results. The left hand side contains original shooting footage, and the right hand side contains the flow visualization encapsulating the change from this shooting keyframe to the next keyframe. For simplicity, we show only the keyframe before the flow is observed and not the keyframe after. 

In the first set of examples, we see a shooter shooting to the left. It is hard for human to detect gunshot visually without listening to the gunshot because the white smoke fades into the background which is also grey and white. However, our optical flow graph clearly shows a blue cloud with very static background. This is a clear signal of a gunshot. 

In the second set of examples, we have a very low-resolution video that any human would find very difficult to tell what is inside the darkness; a few people with practiced eyes or with military training may detect gunshots visually. However, this is very hard, neither people nor gun are visually apparent. When we turn to the flow graph, we see very clear a blot of purple cloud which is a very clear signal of gunshot. 

In the last example, the same level of visual obscurity is observed, the entire video is grey instead of dark, and what makes this video harder is the angle of the camera is from above and far away from the shootings. There are also many people in the video. However, in the flow visualization, we see a very clear and concentrated red dot with a tint of green, signally gunshot clearly. 

\begin{figure}[t]
  \center
  \includegraphics[scale=0.37]{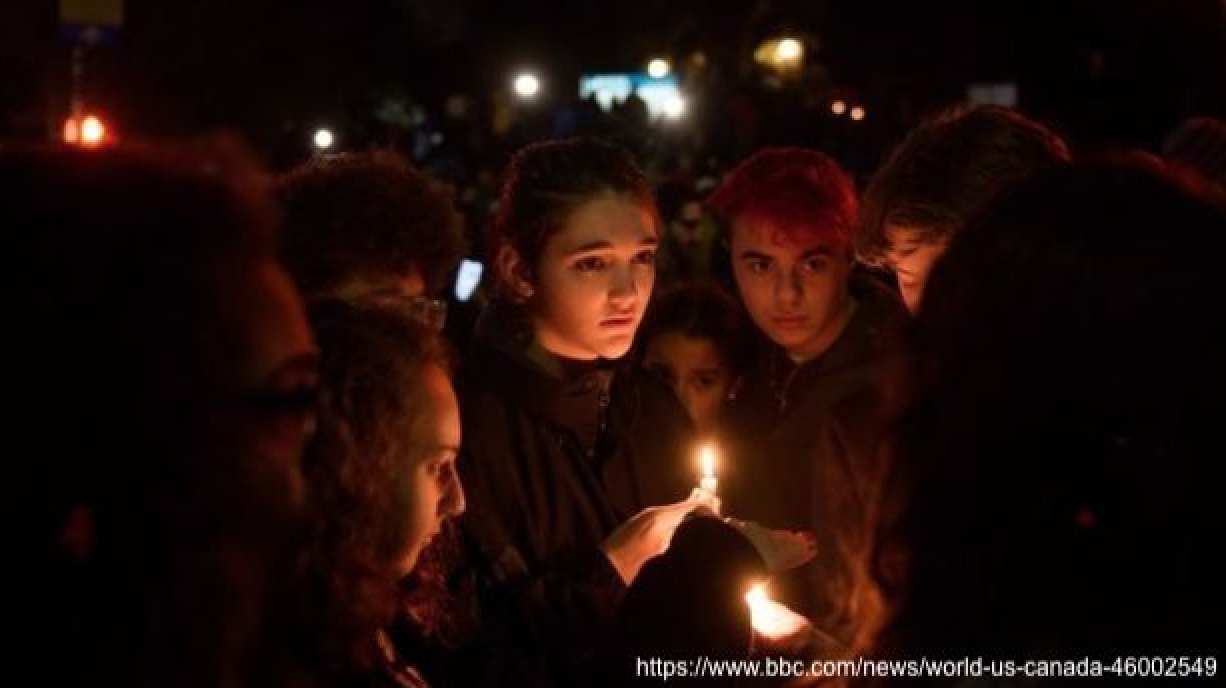}
  \caption{People Gathered to Pray for Victims' Families \cite{PGtPfVF}.}
  \label{fig:prayer}
\end{figure}
\section{Conclusion}
We focus on gun source detection and muzzle head detection in the hope of mitigating gun violence and helping the police to detect the severity of a gun shooting incident as fast as possible. The huge need across the world for protection against gun violence is our main motivation and we as researchers hope to contribute our share to world peace and social stability. 

In our research, we turn our problem formulation to make use of human object detection and gun smoke detection and have interesting results both in bounding the shooter(s) as well as detecting the gun smoke. The muzzle head is found between the gun smoke and the shooter. 

Indeed, when we overlay the optical flow visualization with the human detection output, we see clearly where the muzzle head is. The muzzle head is between the gun smoke on the left and the shooter on the right. We label the shooter, muzzle head and gun smoke as shown in Figure \ref{fig:final}.
Our results for gun smoke detection are much higher than for shooter detection; in Table \hyperref[tbl:final]{2}, our evaluation is done using the 15 videos having visible gunshots, gun cloud detection's success rate is 69.6\% while shooter human detection's success rate is 30.4\%. Since muzzle head detection (21.7\%) relies on both gun cloud detection and shooter detection, the success rate of muzzle head detection has a great potential to improve, and benefits the most if we can improve the shooter human detection rate as shown in Table \hyperref[tbl:final]{2}.  

In the future, researchers may attempt to locate the exact location of the muzzle head in the direction of the gun smoke. The muzzle head should be on the straight line connecting the center of the gun cloud and the shooter's eye. However, though it is relatively easier to detect people, it is hard to locate a person's eyes. Also, though the gun cloud is visible and in most cases round, sometimes it is oval or irregular. Therefore, it may be hard to identify the center of the cloud. But all of these are interesting questions for future research. 

\section*{Acknowledgments}
We would like to thank Carla De Oliveria Viegas, Siddhartha Sharma, Junwei Liang, Ruonan Liu and Ankit Shah for their valuable feedbacks and their generous sharing of ideas, data, models, and feature files. 
%
\clearpage
\bibliographystyle{ieeetr}
\bibliography{bibliography}

\appendix

\end{document}